\pdfoutput=1
\documentclass[11pt,a4paper]{article}
\usepackage[hyperref]{emnlp2020}
\usepackage{times}
\usepackage{inconsolata}

\usepackage{url}
\usepackage{bibunits}

\newif\ifcomment
\commenttrue

\newif\ifdoublespaceme
\doublespacemefalse %
\usepackage{framed}
\usepackage{mdwlist}
\usepackage{latexsym}
\usepackage{nicefrac}
\usepackage{booktabs}
\usepackage{amsfonts}
\usepackage{bold-extra}
\usepackage{amsmath}
\usepackage{dsfont}
\usepackage{amssymb}
\usepackage{bm}
\usepackage{graphicx}
\usepackage{mathtools}
\usepackage{microtype}
\usepackage{multirow}
\usepackage{multicol}
\usepackage{xspace} %
\usepackage{comment}
\usepackage{subfigure}

\usepackage{xcolor}
\usepackage[nomessages]{fp}
\usepackage{tikz}
\usepackage{blindtext}
\usepackage{calc}
\usepackage{pgfplots, pgfplotstable}
\usepackage{filecontents}

\pgfplotstableset{col sep=comma}

\newcommand{\gem}[1]{\mbox{\textsc{gem}}}
\newcommand{\abr}[1]{\textsc{#1}}

\newcommand{\g}{\, | \,}

\newcommand{\hidetext}[1]{}
\newcommand{\ignore}[1]{}

\ifcomment
\newcommand{\todo}[1]{\textcolor{red}{{\bf TODO: #1}}}
\else
\newcommand{\todo}[1]{}
\fi

\ifcomment
\newcommand{\pinaforecomment}[3]{\colorbox{#1}{\parbox{.8\linewidth}{#2: #3}}}
\else
\newcommand{\pinaforecomment}[3]{}
\fi

\newcommand{\smallurl}[1]{ \begin{tiny}\url{#1}\end{tiny}}

\definecolor{lightblue}{HTML}{3cc7ea}
\definecolor{CUgold}{HTML}{CFB87C}
\definecolor{grey}{rgb}{0.95,0.95,0.95}
\definecolor{ceil}{rgb}{0.57, 0.63, 0.81}

\newcommand{\wonderdata}{\textsc{Squall}\xspace} %
\newcommand{\wtqfull}{\textsc{Wiki\-Table\-Questions}\xspace}
\newcommand{\wtq}{\textsc{WTQ}\xspace}
\newcommand{\glove}{\abr{gl}{\small o}\abr{ve}\xspace}
\newcommand{\base}{\abr{seq2seq}\xspace}
\newcommand{\baseline}{\abr{seq2seq}\textsuperscript{+}\xspace}

\newcommand{\baselinebert}{\abr{seq2seq}\textsuperscript{+} w/ \abr{bert}\xspace}

\newcommand{\supattn}{\abr{sup-attn}\xspace}

\newcommand{\alignmodel}{\abr{align}\xspace}
\newcommand{\alignmodelbert}{\abr{align} w/ \abr{bert}\xspace}

\newcommand{\acclf}{$\text{ACC}_{\text{LF}}$\xspace}
\newcommand{\acclfminus}{$\text{ACC}_{\text{LF}}^{-}$\xspace}
\newcommand{\acccol}{$\text{ACC}_{\text{COL}}$\xspace}

\newcommand{\accexe}{$\text{ACC}_{\text{EXE}}$\xspace}
\newcommand{\accskt}{$\text{ACC}_{\text{TEMP}}$\xspace}

\newcommand{\sql}[1]{\texttt{#1}}

\newcommand{\reftab}[1]{Table~\ref{#1}}
\newcommand{\reffig}[1]{Figure~\ref{#1}}
\newcommand{\refsec}[1]{\S\ref{#1}}

\newcommand{\refeqn}[1]{Eq.~\eqref{#1}}

\newlength\maxlen
\newlength\unitlen

\newcommand{\posscite}[1]{\citeauthor{#1}'s \citeyearpar{#1}}

\definecolor{exampleblue}{RGB}{189, 215, 238}
\definecolor{exampletextblue}{RGB}{47, 85, 151}

\makeatletter
\renewcommand\sectionautorefname{\S\@gobble}  

\aclfinalcopy %

\def\aclpaperid{****} %

\title{On the Potential of Lexico-logical Alignments\\for Semantic Parsing to SQL Queries}

\author{Tianze Shi\thanks{~~Equal contribution; listed in alphabetical order.}\\
  Cornell University\\
  {\tt tianze@cs.cornell.edu} \\\And
  Chen Zhao\footnotemark[1]\\
  University of Maryland\\
  {\tt chenz@cs.umd.edu} \\\And
  Jordan Boyd-Graber\\
  University of Maryland\\
  {\tt jbg@umiacs.umd.edu} \\\AND
  Hal Daum{\'e} III\\
  Microsoft Research \& 
  University of Maryland\\
  {\tt me@hal3.name} \\\And
  Lillian Lee\\
  Cornell University\\
  {\tt llee@cs.cornell.edu} \\}

\date{}

\begin{document}
\maketitle

\ifdoublespaceme
  \doublespacing
\fi

\begin{abstract}
Large-scale semantic parsing datasets annotated with logical forms have
enabled major advances in supervised approaches. But can
richer supervision help even more?
To explore the
utility of fine-grained, lexical-level supervision,
we introduce \wonderdata,
a dataset that enriches  $11{,}276$  \wtqfull
 English-language questions with manually created \abr{sql} equivalents plus alignments between \abr{sql} and question fragments.
Our annotation enables new training
possibilities
for encoder-decoder models,
including approaches from machine translation previously precluded by the absence of alignments.
We propose and test two methods:
(1) supervised attention; (2)
adopting
an auxiliary objective of disambiguating
references in the input queries to table columns.
In $5$-fold cross validation,
these strategies improve over strong baselines by $4.4\%$
execution accuracy.
Oracle experiments suggest that
annotated alignments
can support further accuracy gains of
up to $23.9\%$.

\end{abstract}

\begin{bibunit}[acl_natbib_2019]

\newcommand*\circled[1]{\tikz[baseline=(char.base)]{
    \node[shape=circle,draw,scale=0.75,inner sep=1pt] (char) {#1};}}

\begin{figure}[!t]
\centering
\includegraphics[width=0.94\linewidth]{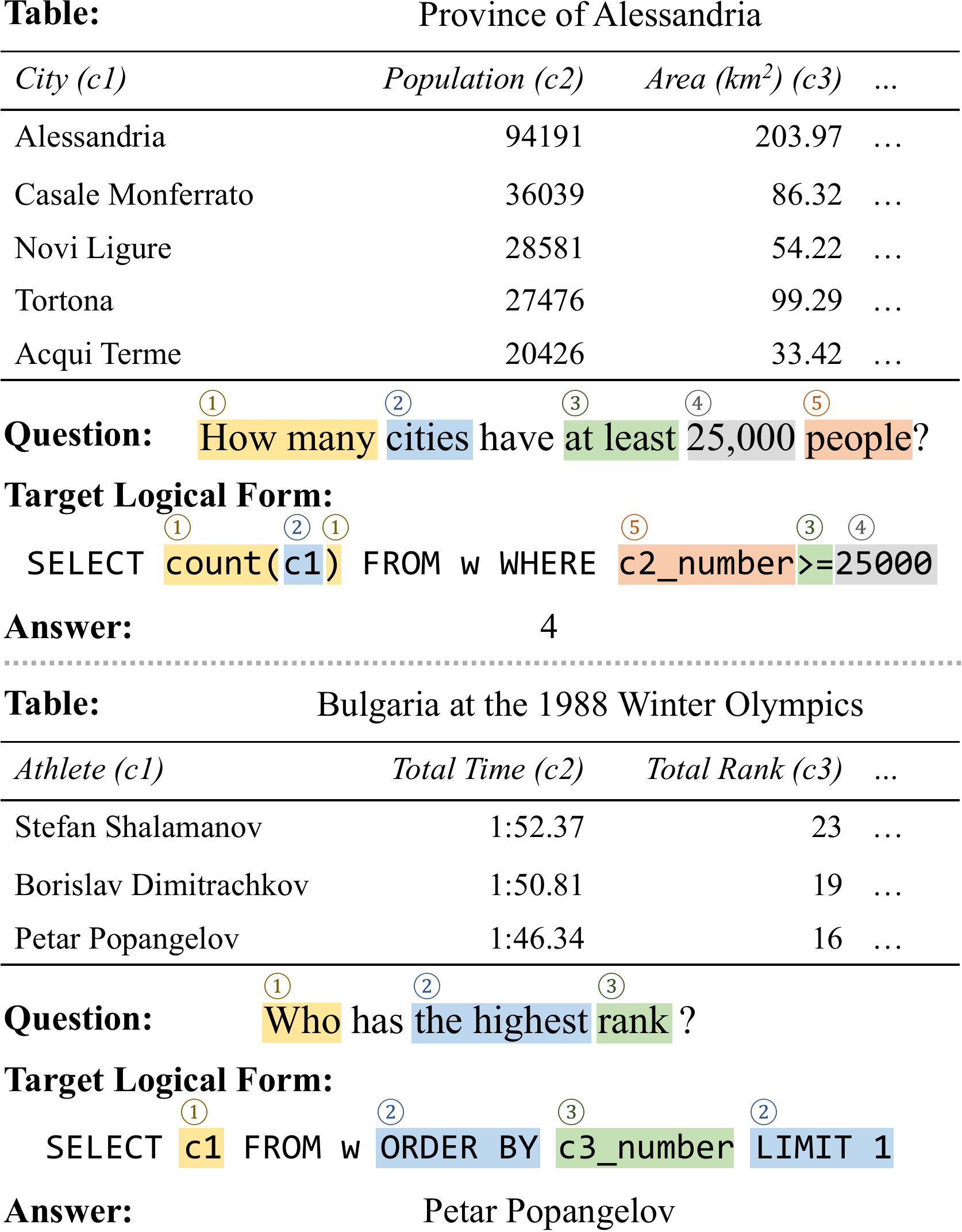}
\caption{Two
examples from \wonderdata.
The table-question-answer triplets come from \wtqfull.
We provide the logical forms as \abr{sql}
plus alignments
between question and
logical form.  In the bottom example, for instance,
``the highest'' $\leftrightarrow$ \sql{ORDER BY} and \sql{LIMIT 1}, as indicated
by both matching highlight color
(\hspace{1pt}\setlength{\fboxsep}{1.5pt}\colorbox{exampleblue}{blue}\hspace{1pt}) and circled-number labels $\left({\bf \textcolor{exampletextblue}{\circled{2}}}\right)$.
}
\label{fig:ex}
\end{figure}

    \begin{tikzpicture}[remember picture,overlay]
    \node[anchor=south,yshift=20pt,xshift=0pt]%
        at (current page.south)
        {\emph{Findings of ACL: EMNLP 2020}};
    \end{tikzpicture}

\section{Introduction}
The availability of large-scale datasets
pairing natural utterances
with  logical forms~\citep[\emph{inter alia}]{dahl1994expanding,wang+15i,zhong2017seq2sql,yu2018spider}
has
enabled significant progress on supervised approaches to semantic parsing~\cite[\emph{inter alia}]{jia2016recombination,xiao+16,dong2016language,dong-lapata-2018-coarse}.
However, the
provision of logical forms alone does not indicate
important fine-grained relationships
between individual words or phrases and logical form tokens.
This is unfortunate because researchers have in fact hypothesized
that the lack of such {\em alignment} information
hampers progress in semantic parsing~\citep[pg. 80]{zhang+19}.

We address this
lack by introducing \wonderdata,\footnote{\wonderdata=``\underline{S}QL+\underline{QU}estion pairs \underline{AL}igned \underline{L}exically''.
}
the first large-scale semantic-parsing dataset with manual lexical-to-logical alignments;
and we investigate the potential accuracy boosts
achievable from such alignments.
The starting point for \wonderdata is \wtqfull \citep[\wtq;][]{pasupat2015compositional},
containing data tables, English questions regarding the tables, and table-based answers.
We manually enrich
the $11{,}276$-instance subset of \wtq's training data that is translatable to \abr{sql}
by providing expert annotations, consisting
not only
of target logical forms in \abr{sql}, but also
labeled alignments between
the input question tokens (e.g., ``how many'') and their corresponding
\abr{sql} fragments
(e.g., \sql{COUNT($\ldots$)}).
\reffig{fig:ex} shows two
\wonderdata instances.
These
new
data enable
training of
encoder-decoder neural models
that incorporates manual alignments.
Consider the bottom example in \reffig{fig:ex}:
A decoder
can benefit from knowing that
\sql{ORDER BY $\ldots$ LIMIT 1}
comes from ``the highest'' (where rank 1 is best);
and an encoder should match ``who'' with the ``athlete'' column
even though the two
strings
have no overlapping tokens.
We implement
these ideas with two
training
strategies:
\begin{enumerate}
\item \emph{Supervised attention} that guides
models to produce attention weights
mimicking human judgments
during both encoding and decoding.
Supervised attention has
improved both alignment and translation
quality in machine translation~\citep{liu+16,mi+16},
but has only been applied in semantic parsing to heuristically generated alignments \citep{rabinovich17}
due to the lack of manual annotations.
\item \emph{Column prediction} that
infers which column in the data table a question fragment refers to.
\end{enumerate}

Using BERT features, our models reach $54.1\%$ execution accuracy on the \wtq test set,
surpassing the previous weakly-supervised state-of-the-art $48.8\%$
(where weak supervision means access to only the answer, not the logical form
of the question).
More germane to the issue of alignment utility,
in $5$-fold cross validation, our
additional
fine-grained supervision improves execution accuracy
by $4.4\%$ over models supervised with only logical
forms;
ablation studies indicate that
mappings between question tokens and columns help the most.
Additionally,
we construct \emph{oracle} models that have access to the full alignments
during test time
to show the unrealized \emph{potential} for our data,
seeing improvements of
up to $23.9\%$ absolute logical form accuracy.
Through annotation-cost and learning-curve analysis,
we conclude that lexical alignments are
cost-effective for training parsers:
lexical alignments take less than half the time to annotate as a logical form
does,
and we can improve execution accuracy by $2.5$
percentage points
by aligning merely $5\%$ of the logical forms in the training set.

Our contributions are threefold:
1) we release a high-quality semantic parsing dataset with manually-annotated logical forms;
2) we label the alignments between the English questions and the corresponding logical forms
to provide additional supervision;
3) we propose two training strategies
that use our alignments to improve
strong base models.
Our dataset and code are publicly available at \url{
  https://www.github.com/tzshi/squall}.

\section{Task: Table-based Semantic Parsing}
\label{sec:task}

Our task is to answer questions about structured tables through semantic parsing to logical forms
(LFs).
Formally, the input $x=(q,T)$ consists of a question $q$ about a table $T$,
and the goal of a semantic parser is to reproduce the 
target LF $y^\star$ 
for $q$ 
(and thus have high \emph{LF accuracy})
or, in a less strict setting, to generate 
any query LF $y'$ that,
when executed against $T$, yields
the correct output $z^\star$ (and thus have high \emph{execution accuracy}).

In a \emph{weakly supervised} setting, training examples 
consist only of input-answer pairs  $(x, z^\star)$.
Recent datasets \citep[\emph{inter alia}]{zhong2017seq2sql,yu2018spider}
provide
enough logical forms, i.e., $(x, y^\star)$ 
training
pairs,
to learn from mappings from $x$ to $y^\star$
in a \emph{supervised} setting.
Unsurprisingly, supervised models
are more accurate
than weakly supervised ones.
However, training supervised models is still challenging:
both $x$ and $y$ are structured,
so models typically generate $y$
in multiple steps,
but the training data cannot reveal
which parts of $x$
generate which parts of $y$
and how they are combined.

Just as adding supervised training improves accuracy over weak supervision,
we explore whether even \emph{finer}-grained supervision further helps.
Since no
large-scale datasets 
furnishing fine-grained supervision
exist (to the best of our knowledge),
we introduce \wonderdata{}.

\section{\wonderdata: Our New Dataset}
\label{sec:data}

\wonderdata
is based on \wtqfull \citep[\wtq;][]{pasupat2015compositional}.
\wtq is a large-scale
question-answering dataset
that contains diverse and challenging crowd-sourced question-answer pairs
over
$2{,}108$
semi-structured Wikipedia tables.
Most of the questions are more than simple table-cell look-ups
and are highly compositional,
a fact that motivated us to study
lexical mappings between questions and logical forms.
We
hand-generate \abr{sql} equivalents of the \wtq queries
and align question tokens with corresponding \abr{sql} query fragments.\footnote{
\abr{sql} is a widely adopted formalism.
Other formalisms including
LambdaDCS \citep{pasupat2015compositional},
have been used on \wtq.
\abr{sql} and LambdaDCS can express roughly the same
percentage of queries:
$81\%$ (our finding) vs. $79\%$ (analysis of a $200$-question sample by \citealp{pasupat-liang2016}).
We leave automatic conversion to and from \abr{sql} to other formalisms and vice versa to future work.
}
We leave lexical alignments of other text-to-\abr{sql} datasets and cross-dataset model generalization \citep{alane2020explor} to future work.

\subsection{Data Annotation}
\label{sec:annotation}

We annotated \wtq's training fold
in three stages:
database construction, \abr{sql }query annotation,
and alignment.
Two expert annotators familiar with \abr{sql} annotated half of the dataset each and then checked each other's
annotations and resolved all conflicts via discussion.
See
\autoref{sec:app-anno} for the annotation guidelines.

\paragraph{Database Construction}

Tables encode semi-structured information.
Each table column usually contains data of the same type: e.g.,
text, numbers, dates, etc.,
as is
typical in relational databases.
While pre-processing the \wtq tables, we considered both basic data types
(e.g., raw text, numbers) and composite types (e.g., lists, binary tuples),
and we suffixed column names with their inferred data types (e.g., \sql{\_number} in \autoref{fig:ex}).
For annotation consistency, all tables were assigned the same name \sql{w}
and columns were given the sequential names \sql{c1},~\sql{c2},\ldots in the database schema,
but we kept the original table headers for feature extraction.
We additionally added a special column \sql{id} to every table
denoting the linear order of its rows.
See \autoref{sec:app-database} for details.

\paragraph{Conversion of Queries to \abr{sql}}

For every question in \wtq's training fold,
we manually created its corresponding \abr{sql} query,
choosing the shortest when there are multiple possibilities,
for instance, we wrote ``\sql{SELECT MAX(c1) FROM w}'' instead of ``\sql{SELECT c1 FROM w ORDER BY c1 DESC LIMIT 1}''.
An exception is that we opted for  less table structure-dependent versions
even if their complexity was higher.
As an example, if the table listed games (\sql{c2}) pre-sorted by date (\sql{c1}),
and the question was ``what is the next game after A?'',
we wrote ``\sql{SELECT c2 FROM w WHERE c1 > (SELECT c1 FROM w WHERE c2 = A) ORDER BY c1 LIMIT 1}''
instead of ``\sql{SELECT c2 FROM w WHERE id = (SELECT id FROM w WHERE c2 = A) + 1}''.
Out of $14{,}149$ questions spanning $1{,}679$ tables,
\wonderdata{} provided \abr{sql} queries for
$11{,}468$ questions, or $81.1\%$.
The remaining $18.9\%$ consisted of questions with non-deterministic answers (e.g., ``show me an example of \ldots''),
questions requiring additional pre-processing
(e.g., looking up a date inside a text-based details column),
and cases where \abr{sql} queries would be insufficiently expressive
(e.g., ``what team has the most consecutive wins?'').

\paragraph{Alignment Annotation}

\begin{table}[t]
  \centering
    \small
    \begin{tabular}{lll}
    \toprule
        & how long & \sql{MAX($\ldots$)} \\
    \midrule
        \multirow{5}{*}{\rotatebox[origin=c]{90}{\parbox[c]{1.4cm}{\centering Frequently aligned to}}}
        & \sql{col}            & the last             \\
        & \sql{MAX(col)-MIN(col)}        & the most \\
        & \sql{col-col}                   & the largest           \\
        & \sql{COUNT(*)} & the highest          \\
        & \sql{COUNT(col)}              & the first        \\
    \bottomrule
    \end{tabular}

    \hfill
  \caption{Examples of frequently-aligned 
  English/LF
  segment pairs,
  illustrating the diversity in the aligned counterparts for the same lexical units.
  \sql{col} is a placeholder for the actual data table column mention.
}

  \label{tab:lexicon-examples}%

\end{table}%

Given a tokenized
question/LF
pair,
the annotators selected and aligned corresponding fragments from the two sides.
The selected tokens did note need to be
contiguous,
but they had to
be units that decompose no further.
For the example in \autoref{fig:ex}, there were three alignment pairs,
where the non-contiguous ``\sql{ORDER BY $\ldots$ LIMIT 1}'' was
treated as an atomic unit and aligned to ``the highest'' in the input.
Additionally, not all tokens on either side needed to be aligned.
For instance, \abr{sql} keywords \sql{SELECT}, \sql{FROM} and question tokens ``what'', ``is'', etc. were mostly
unaligned.
\autoref{tab:lexicon-examples} shows that
the same question phrase was aligned to a range of \abr{sql} expressions, and vice versa.
Overall, $49.8\%$ of
question tokens
were
aligned.
Comparative and superlative question tokens were the most frequently aligned,
while many function words were unaligned;
see \autoref{sec:app-pos-tag}
for part-of-speech distributions of the aligned and unaligned tokens.
Except for the four keywords in the basic structure ``\sql{SELECT $\ldots$ FROM w WHERE $\ldots$}'',
$90.2\%$ of \abr{sql} keywords were aligned.
The rest of the unaligned \abr{sql} tokens include d\sql{=} (alignment ratio
of $18.0\%$), \sql{AND} ($25.5\%$) and column names ($86.1\%$).
The first two cases arose
because equality checks and conjunctions of filtering conditions are
often implicit
in natural language.
\paragraph{Inter-Annotator Agreement and Annotation Cost}
The two annotators'
initial \abr{sql} annotation agreement
in a pilot trial\footnote{
In the pilot study, the annotators
independently labeled questions over the same $50$ tables.
We report the
percentage of cases where one annotator
accepted the other annotator's labels.
} was $70.4\%$ and
after discussion, they agreed on $94.5\%$ of data instances;
similarly, alignment agreement rose from $75.1\%$ to $93.3\%$.
With respect to annotation speed,
an average \abr{sql} query took $33.9$ seconds
to produce
and an additional $15.0$ seconds to enrich with alignments:
the cost of annotating $100$ instances with alignment enrichment was comparable to
that of $144$ instances with only logical forms.

\subsection{Post-processing}
\label{sec:data-post}

Literal values in the \abr{sql} queries such as ``25,000'' in \autoref{fig:ex} and ``star one'' in \autoref{fig:case}
are often directly copied from the input questions.
We thus adapted WikiSQL's~\cite{zhong2017seq2sql} task setting,
where all literal values correspond to spans in the input questions.
We used our alignment to generate gold selection spans,
filtering out instances
where literal values could not be reconstructed through fuzzy match from the gold spans.
After post-processing, \wonderdata contained $11{,}276$ table-question-answer triplets
with logical form and lexical alignment annotations.

\section{(State-of-the-Art)\footnote{
In Appendix \refsec{sec:app-sota}, we show that on \wonderdata,
our base model is competitive with a state-of-the-art system \cite{alane2020explor}
benchmarked on the Spider dataset \cite{yu2018spider}.}
Base Model: Seq2seq with Attention and Copying}
\label{sec:base-model}

Recent state-of-the-art text-to-\abr{sql} models extend the sequence-to-sequence (seq2seq) framework
with attention and copying mechanisms \cite[][\emph{inter alia}]{zhong2017seq2sql,dong2016language,dong-lapata-2018-coarse,alane2020explor}.
We adopt this strong neural paradigm as our base model.
The seq2seq model generates one output token at a time
via
a probability distribution
conditioned on both the input sequence representations
and the partially-generated output sequence:
$
P(y \g \mathbf{x})=\textstyle\prod\nolimits_{i=1}^{|y|}{P(y_i \g \mathbf{y}_{<i},\mathbf{x})},
$
where $\mathbf{x}$ and $\mathbf{y}$ are the feature representations for the input and output sequences,
and $_{<i}$ denotes a prefix.
The last token of $y$ must be a special \texttt{<STOP>} token
that terminates the output generation.
The per-token probability distribution is modeled through Long-Short Term Memory networks (LSTMs, \citealp{hochreiter-schmidhuber97})
and multi-layer perceptrons (MLPs):
\begin{align}
    \label{eqn:hi}
    \mathbf{h}_i&=\text{LSTM}(\mathbf{h}_{i-1},\mathbf{y}_{i-1})\\
    \label{eqn:output}
    P(y_i \g \mathbf{y}_{<i},\mathbf{x})&= \text{softmax}\left(\text{MLP}(\mathbf{h}_i)\right).
\end{align}

\noindent
The training objective is the negative log likelihood of the gold $y^\star$,
defined for each timestep as
\begin{equation*}
    L_i^{\text{seq2seq}}=-\log P(y_i^\star \g \mathbf{y}_{<i}^\star,\mathbf{x}).
\end{equation*}

\paragraph{Question and Table Encoding}
An input $x$ contains a length-$n$ question $q=q_1,\ldots,q_n$
and a table with $m$ columns $c=c_1,\ldots,c_m$.
The input question is represented through a bi-directional LSTM (bi-LSTM) encoder that summarizes information
from both directions within the sequence.
Inputs to the bi-LSTM are concatenations of word embeddings,
character-level bi-LSTM vectors,
part-of-speech embeddings, and named entity type embeddings.
We denote the resulting feature vector associated with $q_i$ as $\mathbf{q}_i$.
For column names, the representation $\mathbf{c}_j$ concatenates the final hidden states of two
LSTMs running in opposite directions that take
the concatenated word embeddings, character encodings, and column data type embeddings
as inputs.
We also experiment with pre-trained BERT feature extractors \citep{devlin+19},
where we feed the BERT model
with the question and the columns as a single sequence delimited by the special \sql{[SEP]} token,
and we take the final-layer representations of the question words and the last token of each column as their representations.

\paragraph{Attention in Encoding}

To enhance feature interaction between the question and the table schema,
for each question word representation
$\mathbf{q}_i$,
we use an attention mechanism to determine its relevant columns and calculate
a linearly-weighted context vector $\mathbf{\widetilde{q}}_i$ as follows:
\begin{align}
\label{eqn:agg}
    \mathbf{\widetilde{q}}_{i} & = \text{Attn}(\mathbf{q}_i, \mathbf{c}) \triangleq \textstyle\sum\nolimits_j{\mathbf{a}_{ij}\mathbf{c}_j},\\
\label{eqn:score}
    \text{where }\mathbf{a}_{ij} &= \text{softmax}_j\left(\mathbf{q}_i^T W^{\text{att}}\mathbf{c}\right).
\end{align}

\noindent
Then we run another bi-LSTM by concatenating the question representation $\mathbf{q}$ and context representation $\mathbf{\widetilde{q}}$ as inputs
to derive a column-sensitive representation $\vec{\mathbf{q}}_i$ for each question word $q_i$.
We apply a similar procedure to get the column representation
$\vec{\mathbf{c}}_j$
for each column.

\paragraph{Attention in Decoding}

During decoding, to allow LSTMs to capture long-distance
dependencies from the input,
we add attention-based features to the recurrent feature definition of
\refeqn{eqn:hi}:
\begin{align}
    \mathbf{v}_{i} &= \text{Attn}(\mathbf{h}_{i}, \vec{\mathbf{q}})\\
    \mathbf{h}_{i} &= \text{LSTM}(\mathbf{h}_{i-1},\left[\mathbf{v}_{i-1};\mathbf{y}_{i-1}\right]).
\end{align}

\paragraph{\abr{sql} Token Prediction with Copying Mechanism}

Since each output token can be an \abr{sql} keyword, a column name or a literal value,
we factor
the probability defined in
\refeqn{eqn:output} into two components:
one that decides the type $t_i\in\{\texttt{KEY},\texttt{COL},\texttt{STR}\}$ of $y_i$:
\begin{equation*}
    P(t_i \g \mathbf{y}_{<i},\mathbf{x})=\text{softmax}\left(\text{MLP}^{\text{type}}(\mathbf{h}_i)\right),
\end{equation*}
and another that predicts the token conditioned on the type $t_i$.
For token type \texttt{KEY}, we predict the keyword token with another MLP:
\begin{equation*}
    P(y_i \g \mathbf{y}_{<i},\mathbf{x},t_i=\texttt{KEY})=\text{softmax}\left(\text{MLP}^{\texttt{KEY}}(\mathbf{h}_i)\right).
\end{equation*}
For \texttt{COL} and \texttt{STR} tokens,
the model selects
directly from the input
column names $c$
or question $q$
via a copying mechanism.
We define a probability distribution with softmax-normalized bilinear scores:
\begin{align*}
    &P(y_i=c_j \g \mathbf{y}_{<i},\mathbf{x},t_i=\texttt{COL})=\text{softmax}_j(\mathbf{s}_{i\boldsymbol{\cdot}}), \\
    &\quad\textrm{where }\mathbf{s}_{ij}=\mathbf{h}_i^\top W^{\texttt{COL}} \mathbf{c}_j.
\end{align*}
Similarly, we define literal string copying from $q$ with
another bilinear scoring matrix $W^{\texttt{STR}}$.

\section{Using Alignments in Model Training}
\label{sec:model}

The model design in \autoref{sec:base-model} includes many latent interactions
within and across the encoder and the decoder.
We now describe how
our manual alignments can enable direct supervision on such previously latent interactions.
Our alignments can be used
as supervision for the necessary
attention weights (\autoref{sec:sup-attn}).
In an \emph{oracle experiment} where we
replace induced attention with manual alignments,
the jump in logical form accuracy shows \emph{alignments are valuable}, if only the 
{models could reproduce them} (\autoref{sec:oracle-attn}).
Moreover, alignments enable
a column-prediction auxiliary task (\autoref{sec:model-align}).

The loss function $L$ of our full model is a linear combination of the loss terms of the seq2seq model, supervised attention, and column prediction:
\begin{equation*}
    L = L^{\text{seq2seq}} + \lambda^{\text{att}} L^{\text{att}} + \lambda^{\text{CP}} L^{\text{CP}},
\end{equation*}
where we define $L^{\text{att}}$ and $L^{\text{CP}}$ below.

\subsection{Supervised Attention}
\label{sec:sup-attn}

Our annotated lexical alignments resemble our base model's attention mechanisms.
At the encoding stage, question tokens and the relevant columns are aligned
(e.g., ``who'' $\leftrightarrow$ column ``athlete'')
which should induce
higher weights in both question-to-column and column-to-question attention
(\refeqn{eqn:agg} and \refeqn{eqn:score});
similarly, for decoding,
annotation reflects
which question words are most relevant to the current output token.
Inspired by improvements from supervised attention in machine translation~\citep{liu+16,mi+16},
we
train
the base model's attention mechanisms
to
minimize the Euclidean distance\footnote{
See \autoref{sec:app-abl-loss} for experiments with other distances.
}
between the human-annotated alignment vector $\mathbf{a}^\star$
and the model-generated attention vector $\mathbf{a}$:
\begin{equation*}
L^{\text{att}}=
\frac{1}{2}\lVert\mathbf{a} - \mathbf{a}^\star\rVert^2.
\end{equation*}
The vector $\mathbf{a}^\star$ is a one-hot vector when the annotation aligns to a single element,
or $\mathbf{a}^\star$
represents a uniform distribution over the subset
in cases where the annotation aligns multiple elements.
\subsection{Oracle Experiments with Manual Alignments}
\label{sec:oracle-attn}

\begin{table}[t]
  \centering
  \small
  \begin{tabular}{lcc}
    \toprule
      Attention type & \acclf{} (Dev) & $\Delta$ \\
    \midrule
      \emph{Induced attention}   & $37.8\pm0.6$ \\
    \midrule
    \emph{Oracle attention} \\
      \quad Encoder only & $51.5\pm1.4$ & $+13.7$ \\
      \quad Decoder only & $49.4\pm0.9$ & $+11.6$ \\
      \quad Encoder + decoder & $61.7\pm0.4$ & $+23.9$ \\
    \bottomrule
  \end{tabular}
    \caption{Oracle experiment LF-accuracy results over five dev sets from random splits, where attention weights are replaced by manual alignments.
    \emph{Induced attention} refers to the base model (\autoref{sec:base-model}).}
  \label{tb:oracle}
\end{table}

To present the potential of alignment annotations for models with supervised attention,
we first assume a model that can flawlessly reproduce our annotations within
the base model.
During training and inference,
we feed the
true alignment
vectors in place of the attention weights to the encoder and/or decoder.
\autoref{tb:oracle} shows the resultant logical form accuracies.
Access to oracle alignments provides up to $23.9\%$ absolute higher accuracy over the base model.
This wide gap suggests the high potential for training models with our lexical alignments.
\subsection{Column Prediction}
\label{sec:model-align}
\citet{wang-etal-2019-learning} show the importance of inferring token-column correspondence in a weakly-supervised setting; \wonderdata enables full supervision for
an auxiliary task that directly predicts the corresponding column $c_j$ for
each question token $q_i$.
We model this auxiliary prediction as:
\begin{align*}
    &\mathbf{s}_{ij}=\mathbf{q}_i^\top W^{\text{CP}} \mathbf{c}_j\\
    &P(q_i \text{ matches } c_j \g q_i)=\text{softmax}_j(\mathbf{s}_{i\boldsymbol{\cdot}}).
\end{align*}
For the corresponding loss $L^{\text{CP}}$ over tokens that match columns,
we use cross-entropy.

\paragraph{Exact-match Features: An Unsupervised Alternative}
A heuristic-based, 
albeit lower-coverage, alternative to manual 
alignment is to %
use
questions' mentions %
of column names.
Thus, we use automatically-generated exact-match features in our baseline models for comparison in our experiments.
For question encoders,
we include two embeddings derived from binary exact-match features:
indicators of whether the token appears in (1) any of the column headers
and (2) any of the table cells.
Similarly, for the column encoders,
we also include an exact-match feature of whether the column name
appears in the question.

\section{Experiments}
\label{sec:experiment}

\begin{table}[t]
  \centering
  \small
  \begin{tabular}{lc}
    \toprule
      Model & \accexe (Test) \\
    \midrule
      \multicolumn{2}{l}{\emph{Prior work} (all necessarily are weakly supervised)} \\
      Single model & $34.2$--$44.5$ \\
      Single model (w/ \abr{bert}) & $48.8$ \\
      Ensemble     & $37.7$--$46.9$ \\
    \midrule
      \multicolumn{2}{l}{\emph{This paper} (strongly supervised for the first time)} \\
      Single model (\alignmodel)     &  $49.7\pm 0.4$  \\
      Single model (\alignmodelbert) &  $54.1\pm 0.2$  \\
      Ensemble (\alignmodel)     &  $53.1$  \\
      Ensemble (\alignmodelbert) &  $57.2$  \\
    \bottomrule
  \end{tabular}
  \caption{
      \wtq test set execution accuracies ($\%$).
      The accuracy ranges for prior work are aggregated over
      \citet{pasupat2015compositional},
      \citet{neelakantan2016learning},
      \citet{krishnamurthy2017neural},
      \citet{zhang+17},
      \citet{haug2018neural},
      \citet{liang+18},
      \citet{dasigi+19},
      \citet{agarwal+19},
      \citet{wang-etal-2019-learning},
      and \citet{herzig+20}.
      Unsurprisingly, our models trained on \wonderdata{}
      surpass weakly-supervised previous work.
    }
  \label{tb:test-res}
\end{table}

\paragraph{Setup}
We randomly shuffle the tables in \wonderdata{} and divide them into five splits.
For each setting, we report the average logical form accuracy \acclf{} 
(output LF exactly matches the target LF)
and
execution accuracy \accexe{} 
(output LF may not match the target LF, but its execution yields the gold-standard answer)
as well as the standard deviation of five models,
each trained with four of the splits as its training set and the other split as its dev set.
We denote the base model from \autoref{sec:base-model} as \base
and our model trained with both proposed training strategies in \autoref{sec:model} as \alignmodel.
The main baseline model we compare with, \baseline,
is the base model enhanced with the automatically-derived exact-match features (\autoref{sec:model-align}).
See Appendix \autoref{sec:app-impl} for model implementation details.

\ignore{
\paragraph{Methods}
We compare the baseline seq2seq model with our methods using annotated alignments.
\begin{itemize*}
    \item Base model (\base{}): Sequence-to-Sequence model with attention and
    	copying mechanism as described in \refsec{sec:base-model}, without alignment annotations.
    \item  Alignments for encoder (\alignmodel{}): Our proposed model incorporates column alignments
    into the base model's encoder with and auxiliary task (\refsec{sec:sup-attn} and \refsec{sec:model-align}).

\end{itemize*}
}

\paragraph{WTQ Test Results}

\autoref{tb:test-res} presents the WTQ test-set \accexe{} of \alignmodel{} compared with previous models.
Unsurprisingly,
\wonderdata{}'s supervision allows
our models to surpass weakly supervised models.
Single models trained with BERT feature extractors exceed prior state-of-the-art by $5.3\%$.
However, our main scientific interest is not these numbers per se,
but how beneficial additional lexical supervision is.

\begin{table}[t]
  \centering
  \small
    \begin{tabular}{@{\hspace{4pt}}l@{\hspace{3.5pt}}@{\hspace{3.5pt}}c@{\hspace{3.5pt}}@{\hspace{3.5pt}}c@{\hspace{3.5pt}}@{\hspace{3.5pt}}c@{\hspace{4pt}}}
    \toprule
        \multirow{2}{*}{Model} & \multicolumn{2}{c@{\hspace{3.5pt}}}{Dev} & \multicolumn{1}{c}{Test}\\
          & \acclf & \accexe  &\accexe \\
    \midrule
    \baseline   &$37.8\pm0.6$&$56.9\pm0.7$ & $46.6\pm0.5$\\
    \alignmodel & $42.2\pm1.5$&$61.3\pm0.8$ & $49.7\pm0.4$\\
      \midrule
    \baselinebert &$44.7\pm2.1$&$63.8\pm1.1$ & $51.8\pm0.4$ \\
    \alignmodelbert & $47.2\pm1.2$ & $66.5\pm1.2$ & $54.1\pm0.2$ \\

    \bottomrule
  \end{tabular}
    \caption{
        Logical form (\acclf) and execution (\accexe) accuracies ($\%$)
        on dev and test sets,
        showing the utility of learning from lexical supervisions.
    }
  \label{tb:res}

\end{table}

\paragraph{Effect of Alignment Annotations}
To examine the utility of lexical alignments as a finer-grained type of supervision,
we compare \alignmodel with \baseline in \autoref{tb:res}.
Both have access to logical form supervision,
but \alignmodel additionally uses lexical alignments
during training.
\alignmodel{} improves \base{} by $2.3\%$ with BERT and $3.1\%$ without,
showing that
lexical alignment annotation is more beneficial than
automatically-derived exact-match column reference features.\footnote{
Test set accuracies are lower than on the dev set
because the \wtq test set includes questions unanswerable by \abr{sql}.
}

\paragraph{Effect of Individual Strategies}
\autoref{tb:ablation} compares model variations.
We add each individual training strategy into the baseline \baseline model
and ablate components from the \alignmodel model.
Each component contributes to increased accuracies compared with \baseline.
The effects range from $+1.3\%$ \accexe{} with column prediction to $+3.8\%$ \accexe{} with supervised encoder attention.
Supervised encoder attention is the single most effective strategy:
including it produces the highest gains and ablating it  the largest drop.
The exact-match column reference features are essential to the baseline model:
\base without those features has $8.1\%$ lower \accexe.
Nonetheless, supervised encoder attention and column prediction are still effective
on top of the exact-match features.
Yet, \alignmodel's accuracy is still far below that of the oracle models;
we hope
\wonderdata{} can inspire future work
to 
take better advantage of its rich supervision.

\newcommand\databar[5][orange!30]{%
    \FPeval\result{round((#4-#2)/(#3-#2):4)}%
  \rlap{\textcolor{#1}{\hspace*{\dimexpr-\tabcolsep+.5\arrayrulewidth}%
        \rule[-0.2\ht\strutbox]{\result\maxlen}{1.2\ht\strutbox}}}%
  \makebox[\dimexpr\maxlen-2\tabcolsep+\arrayrulewidth][c]{#5}}
\settowidth\maxlen{$00.00\pm0.0$}
\addtolength\maxlen{\dimexpr0\tabcolsep-\arrayrulewidth}

\newcommand\ablatebar[2]{\multicolumn{1}{|c}{\databar{30}{70}{#1}{$#1\pm#2$}}}
\setlength{\unitlen}{0cm}
\addtolength\unitlen{\dimexpr\maxlen/(70-30)}

\begin{table}[t]
  \centering
  \footnotesize
  \begin{tabular}{lcc}
    \toprule
    \multirow{2}{*}{Component} & \multicolumn{2}{c}{Dev}\\
    & \acclf & \accexe \\
    \midrule
    \base  & \ablatebar{31.0}{0.7}& $48.8\pm0.8$\\
    \midrule
      \baseline                  & \ablatebar{37.8}{0.6}  & $56.9\pm0.7$ \\
      + Supervised decoder attn.   & \ablatebar{39.4}{1.1}  & $58.6\pm1.3$ \\
      + Supervised encoder attn.   & \ablatebar{41.3}{1.7}  & $60.7\pm0.7$ \\
      + Column prediction    & \ablatebar{38.6}{0.5}  & $58.2\pm0.8$ \\
      \midrule
      \alignmodel{}          & \ablatebar{42.2}{1.5}  & $61.3\pm0.8$  \\
      - Supervised decoder attn.    & \ablatebar{41.6}{1.8}  & $61.1\pm1.3$ \\
      - Supervised encoder attn.    & \ablatebar{39.6}{0.6}  & $58.7\pm0.8$ \\
      - Column prediction    & \ablatebar{41.8}{1.6}  & $60.9\pm0.8$ \\
      - Exact-match features & \ablatebar{39.5}{1.1}  & $58.8\pm0.7$ \\
    \midrule
      Oracle attention       & \ablatebar{61.7}{0.4}  & --  \\

      \bottomrule
      \addlinespace[-0.5pt]
      \multicolumn{1}{c}{}
    &
      \multicolumn{1}{c}{
    \makebox[\dimexpr\maxlen-2\tabcolsep+\arrayrulewidth][c]{
        \begin{tikzpicture}[trim left]
            \draw[->] (0,0) -- (\maxlen,0);
            \foreach \x in {35,45,...,65}
                \draw[thin] ({(\x-30)*\unitlen},0) -- ({(\x-30)*\unitlen},-0.05);
            \foreach \x in {30,40,50,60}
                \draw[thin] ({(\x-30)*\unitlen},0) -- ({(\x-30)*\unitlen},-0.1);
            \foreach \x in {30,40,50,60}
                \node at ({(\x-30)/22}, {0-0.1}) [below,scale=0.7] {{\footnotesize \x}};
        \end{tikzpicture}
    }}
      &\\
  \end{tabular}
    \vspace{-13pt}
    \caption{
    Dev logical form (\acclf) and execution (\accexe) accuracies for different model variations (w/o \abr{bert}).
    The superimposed bar chart provides a visual presentation of \acclf.
    Each \alignmodel component contributes to increased accuracies
    compared with \baseline,
    while the oracle attention model demonstrates
    the unrealized potential of the alignments.
    }
  \label{tb:ablation}
\end{table}

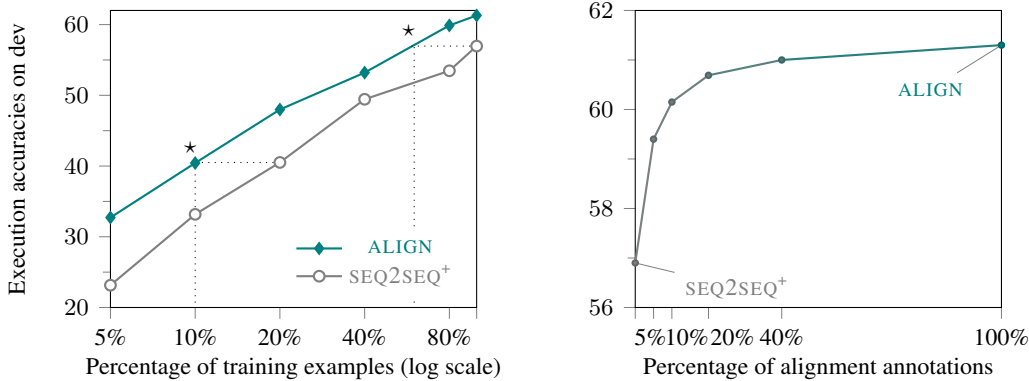
\begin{figure*}[t]
\centering
\begin{tikzpicture}[]
    \begin{axis}[
        width=0.4*\textwidth,
        grid=none,
        xlabel={Percentage of training examples (log scale)},
        ylabel={Execution accuracies on dev},
        xmode=log,
        xmin=5,
        xmax=100,
        xtick={5,10,20,40,80,100},
        xticklabels={5\%,10\%,20\%,40\%,80\%},
        ymin=20,
        ymax=62,
        minor y tick num=1,
        xtick align=outside,
        ytick align=outside,
        xtick pos=left,
        ytick pos=left,
        legend pos=south east,
        legend style={draw=none, font=\small, row sep=-1pt},
        yticklabel style = {font=\small},
        xticklabel style = {font=\small},
        xlabel style = {font=\small},
        ylabel style = {font=\small},
    ]

    \addplot [teal,mark=diamond*,thick,mark size=2pt] table [x=perc, y=align] {learning-curve.dat};
    \addplot [gray,mark=*,mark options={fill=white},thick,mark size=2pt] table [x=perc, y=seq2seq] {learning-curve.dat};
    \draw [dotted,-] (axis cs:20,40.5) -- (axis cs:10,40.5);
    \node [font=\small] at (axis cs:9.6,42.7) {$\star$};
    \draw [dotted,-] (axis cs:10,0) -- (axis cs:10,40.5);
    \draw [dotted,-] (axis cs:100,56.96) -- (axis cs:57,56.96) node[near end,above left,font=\small]{$\star$};
    \draw [dotted,-] (axis cs:60,0) -- (axis cs:60,56.96);
    \legend{\textcolor{teal}{\alignmodel},\textcolor{gray}{\baseline}};
    \end{axis}
\end{tikzpicture}
\hspace{20pt}
\begin{tikzpicture}[]
    \begin{axis}[
        width=0.4*\textwidth,
        grid=none,
        xlabel={Percentage of alignment annotations},
        xmin=0,
        xmax=100,
        xtick={0,5,10,20,40,100},
        xticklabels={,5\%~~,~~~~~10\%,~~~~~~~~20\%,40\%,100\%},
        ymin=56,
        ymax=62,
        minor y tick num=1,
        xtick align=outside,
        ytick align=outside,
        xtick pos=left,
        ytick pos=left,
        legend pos=south east,
        legend style={draw=none, font=\small, row sep=-1pt},
        yticklabel style = {font=\small},
        xticklabel style = {font=\small},
        xlabel style = {font=\small},
        ylabel style = {font=\small},
        point meta=x,
    ]

    \addplot [scatter,mark=*,thick,mark size=1pt,mesh,colormap={}{color(0cm)=(gray); color(1cm)=(teal);}] table [x=perc, y=accexe] {learning-curve-align.dat};
    \node [coordinate,pin={-5:{\small\textcolor{gray}{\baseline}}}] at (axis cs:  0, 56.9) {};
    \node [coordinate,pin={-135:{\small\textcolor{teal}{\alignmodel}}}] at (axis cs:  100, 61.3) {};
    \end{axis}
\end{tikzpicture}

    \caption{
    (Left) the $\star$ markers on the learning curves illustrate that
    \alignmodel uses roughly half the amount of training data
    to achieve similar \accexe as \baseline.
    (Right) annotating just $5\%$ of the logical forms with alignments
    yields \emph{half} of the accuracy improvement of \alignmodel.
    }

\label{fig:learning-curve}
\end{figure*}

\paragraph{Effect of Annotation Availability: Are Lexical Alignments Worth It?}
The lefthand side  of \autoref{fig:learning-curve} plots 
\baseline's and \alignmodel's learning curves.
For each of for \baseline's accuracy levels, \alignmodel reaches a similar level but at
the much ``cheaper'' training cost of about half as many training examples.
Moreover, 
the righthand side of \autoref{fig:learning-curve} shows what happens if \alignmodel
has access to all the training logical forms, but only a percentage of the
accompanying alignments.
Surprisingly,
more than half of the accuracy improvement
comes from as little as $5\%$ of the alignment annotations.
Because the cost of aligning
an example is less than half of that for writing a logical form (\autoref{sec:annotation}),
we conclude that annotating lexical alignments is a cost-effective approach
on a fixed budget.

\paragraph{Where Do Our Models Improve the Most?}

According to \autoref{tb:temp-result}, 
\alignmodel produces the highest gains with respect to \baseline on the subtask of column selection ($+4.9\%$),
compared with a $+2.0\%$ improvement on generating correct \abr{sql} templates.
The gain is larger on complex \abr{sql} templates (i.e., those with more 
aggregation functions and nested queries).\footnote{For example, on template \sql{SELECT COUNT(col) FROM w}, the \acccol is 59.4 (\alignmodel) vs. 48.9 (\baseline).
See Appendix \refsec{sec:app-temp} for detailed result breakdowns.
}
which demonstrates the effectiveness of
reinforcing question-column correspondence
through supervised attention and a column prediction auxiliary task.

\paragraph{Do Our Models Generalize Better to Unseen Query Templates?}
We follow \citet{finegan+2018} and consider a challenging evaluation setting
where the models are tested on unseen \abr{sql} query templates.
In \reftab{tb:temp-generalize}, \alignmodel shows an even larger margin compared with \baseline in this setting,
suggesting that lexical alignment supervision
benefits model robustness.
See \autoref{sec:app-generalization} for detailed results.

\begin{table}[t]
  \centering
  \small
  \begin{tabular}{lccc}
    \toprule
    & \acclf  & \accskt & \acccol  \\
    \midrule
  \baseline & $37.8$ & $64.7$ & $39.6$ \\
  \alignmodel & $42.2$ & $66.7$ & $44.5$ \\
  (delta)     & ($+4.4$) & ($+2.0$) & $(+4.9)$ \\
  \bottomrule
  \end{tabular}
    \caption{
      Dev logical form (\acclf), template (\accskt) and column (\acccol) accuracies.
      Parenthetical numbers are deltas with respect to the baseline.
      \alignmodel 
      improves \acccol the most.
    }
  \label{tb:temp-result}

\end{table}

\begin{table}[t]
    \centering
    \small
      \begin{tabular}{lcc}
        \toprule
       \emph{Unseen Templates} & \acclf  & \accexe \\
        \midrule
      \baseline & $15.5$ & $44.8$ \\
      \alignmodel & $26.1$ & $57.3$ \\
      \bottomrule
      \end{tabular}
      \caption{
        Model accuracies in a generalization setting:
        we exclude an \abr{sql} template from training, and evaluate on that unseen template.
        Shown are macro-averages over the $10$ most frequent templates.
        \alignmodel is more accurate than \baseline by a large margin.
      }
    \label{tb:temp-generalize}

  \end{table}

\begin{table}[t]
    \centering
    \small
    \begin{tabular}{l|c@{\hspace{8pt}}c@{\hspace{8pt}}c|c@{\hspace{8pt}}c@{\hspace{8pt}}c}

      \midrule
      \multirow{2}{*}{Model}
                      & \multicolumn{3}{c|}{Recall} & \multicolumn{3}{c}{Entropy} \\
                      & q2c    & c2q     & d2q    & q2c    & c2q    & d2q    \\
       \midrule
       \baseline      & $26.1$ & ~~$4.8$ & $33.2$ & $0.31$ & $0.16$ & $1.24$ \\
       + Sup. enc.    & $64.8$ & $66.0$  & $35.6$ & $1.57$ & $1.95$ & $1.10$ \\
       + Sup. dec.    & $55.5$ & ~~$3.9$ & $86.6$ & $0.44$ & $0.24$ & $0.99$ \\
       \alignmodel{}  & $65.4$ & $65.9$  & $86.2$ & $1.56$ & $1.94$ & $1.00$ \\

      \bottomrule
    \end{tabular}
      \caption{
        Recall against hand-annotated alignments and average entropy of the attention distributions
        in the question-to-column (q2c), column-to-question (c2q) and decoder-to-question (d2q) modules,
        comparing models trained with supervised encoder/decoder attention, none (\baseline), or both strategies (\alignmodel).
      }
    \label{tb:align-res}
  \end{table}

\paragraph{Are the Induced Attention Weights Similar to Manual Alignments?}
\autoref{tb:align-res}
quantitatively compares the attention distributions.
The models trained with and without supervised attention
have very different attention patterns:
without explicit supervision, the models focus on a few items (low entropy values),
but those items are usually unlike manually-derived alignments (low recall).
Interestingly, the supervised decoder attention encourages the model to induce
question-to-column (q2c) attention that seems similar to human alignment judgments.  
This is an arguably surprising benefit, since the supervised decoder was not
trained with q2c supervision, and so one might have expected it to perform similarly
to \baseline.
However, one needs to be careful in interpreting these results,
as machine-induced attention distributions are not
intended for direct human interpretation~\citep{jain-wallace19,wiegreffe-pinter19}.

\paragraph{Qualitative Analysis}

\begin{figure}[!t]
\centering
\includegraphics[width=\linewidth]{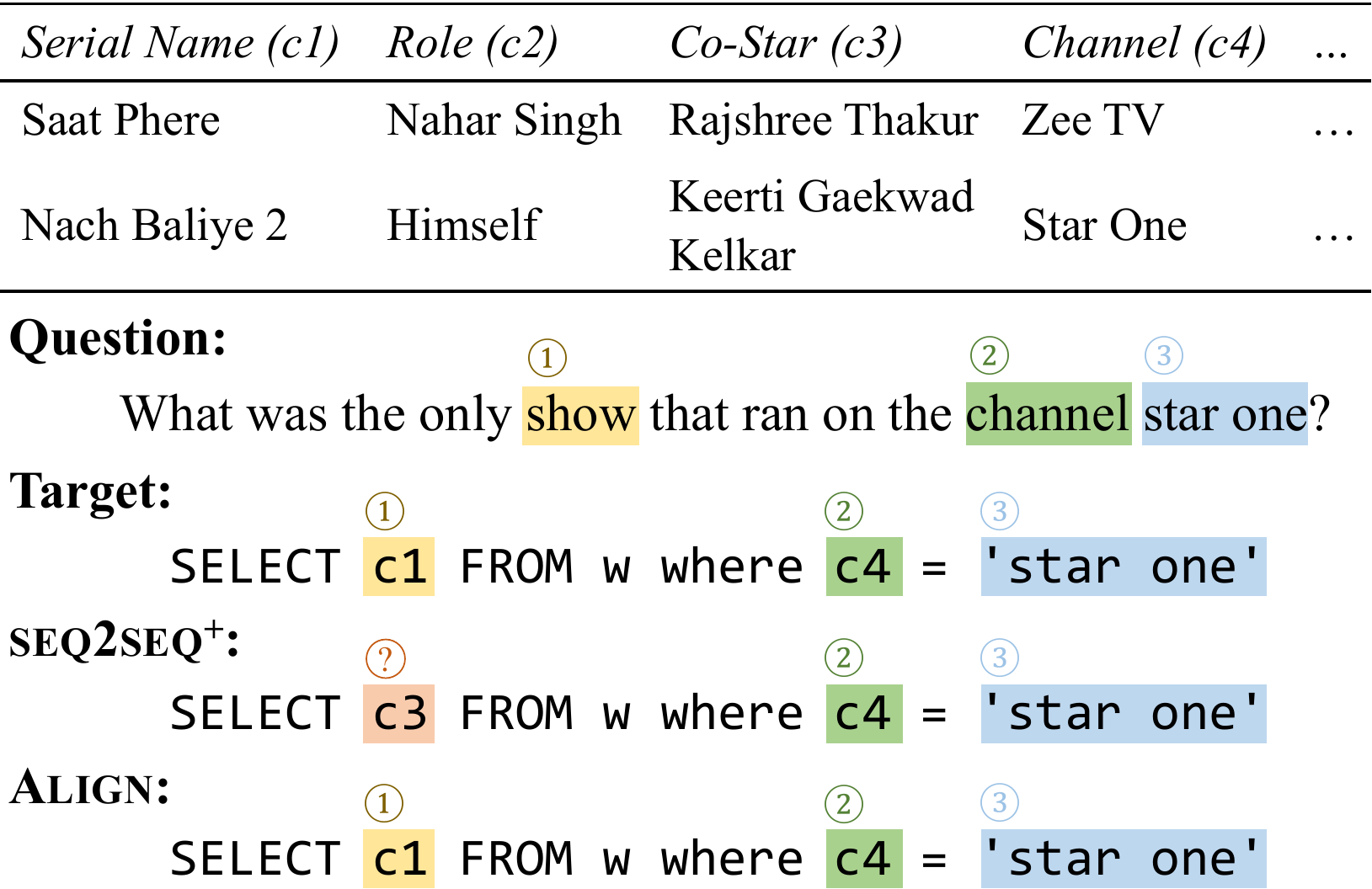}
\caption{An example with \baseline{} and \alignmodel{} predictions.
\baseline{} selects an incorrect column.
}
\label{fig:case}
\end{figure}

Our additional supervision helps when the question
has little textual overlap with the referred columns.
\autoref{fig:case} shows an example.
With finer-grained supervision, \alignmodel{} learns the column ``Serial Name''
corresponds to the question word ``show'',
but \baseline{} selects the wrong column ``Co-Star''.

\ignore{
\paragraph{SQL Complexities}

\reffig{fig:complexity} shows the execution accuracies with different lengths of target \abr{sql} queries.
The improvement of \supattn{} ove the \base{} model %
is mostly uniform across all lengths of queries.
}

\section{Related Work}
\label{sec:related}

\paragraph{Attention and Alignments}
Explicit supervision for attention mechanisms \cite{bahdanau+14}
is helpful for
many tasks, including machine translation \citep{liu+16,mi+16},
image captioning \citep{liu2017attention}, and visual question answering \citep{gan2017vqs}.
For semantic parsing, \citet{rabinovich17}
improve code generation
with exact string-match heuristics to
provide supervision for attention.
\citet{wang-etal-2019-learning} argue that structured alignment is crucial to text-to-\abr{sql} models
and they induce latent alignments in a weakly-supervised setting.
In contrast, we take a fully-supervised approach and train models with manual alignments.
\paragraph{Lexical Focus and Semantic Parsing}%
Our lexical alignment annotations
are similar to semantic lexicons in
lexicalized-grammar-based semantic parsing \cite{Zettlemoyer:2005:LMS:3020336.3020416,zettlemoyer2007online,kwiatkowski2010inducing,krishnamurthy2012weakly,artzi2013weakly}.
Those lexicons are usually well-typed to support semantic composition.
It is an interesting future direction to explore how to model analogous compositional aspects
with our type-flexible alignments
through, for example, syntax-based alignment~\citep{zhang-gildea04}.

\paragraph{Annotator Rationales}

A related direction to enriching annotations is supplying
annotator rationales \citep{zaidan+07},
i.e., evidence supporting the annotations
in addition to the final labels.
Many recent datasets on machine reading comprehension
and question answering,
such as HotpotQA \citep{yang+18b} and CoQA \citep{reddy+19},
include such 
intermediate annotations at dataset release.
\citet{dua+20} show that these annotator rationales
improve model accuracy for a given annotation budget
on machine reading comprehension.
The alignments we provide could, at a stretch, be considered a type of rationale
for the output \abr{SQL} annotation.

\paragraph{Text-to-\abr{sql} Datasets}
There is growing interest in both the database and NLP communities in text-to-\abr{sql} applications.
Widely-used domain-specific datasets include
ATIS \citep{price1990,dahl1994expanding},
GeoQuery \citep{zelle1996learning,popescu2003},
Restaurants \citep{tang-mooney00,popescu2003}, and
Scholar \citep{iyer+17}.
WikiSQL~\cite{zhong2017seq2sql} is among the first large-scale datasets with
question-logical form pairs querying
a wide range of
data tables extracted from Wikipedia, but
WikiSQL's logical forms are generated from a limited set of templates.
In contrast,
\wtq questions are authored by humans under
no specific constraints,
and as a result \wtq includes
more diverse
semantics and logical operations.
The family of Spider datasets~\cite{yu2018spider, yu2019cosql, yu2019sparc}
contain queries 
even more complex than in \wtq, 
including a higher percentage of nested queries and multiple table joins.
We leave extensions of lexical alignments to Spider's complex-structure queries
to future work.

\section{Conclusion}
\label{sec:conclu}

We introduce \wonderdata, the first large-scale semantic parsing dataset
with both hand-produced target logical forms
and manually-derived lexical alignments between questions and \abr{sql} queries.
Our dataset enables finer-grained supervision than existing datasets have previously supported.
We incorporate
the alignments into encoder-decoder-based neural models
through supervised attention and an auxiliary task of column prediction.
Experiments confirm our intuition that finer-grained supervision is helpful to model training.
Our oracle studies
also show that there is large unrealized further potential for our annotations.
Thus,
it remains an exciting challenge for future research to
use our lexical alignment annotations more effectively.

Our annotation cost analysis shows that collecting additional lexical alignments is
more cost-effective for improving model accuracy than having only logical forms.
We hope that our findings will help future dataset design decisions
and extensions of other existing datasets.
One potential future direction is
to further investigate the utility of lexical alignments
in a cross-dataset/domain evaluation setting \citep{alane2020explor}.

\section*{Acknowledgments}

We thank the members of UMD CLIP, Xilun Chen, Jack Hessel, Thomas M\"{u}ller, Ana Smith, and the anonymous reviewers
and meta-reviewer
for their suggestions and comments.
TS was supported by a Bloomberg Data Science Ph.D. Fellowship.
CZ and JBG are supported by the Defense Advanced Research Projects Agency (DARPA) and Air
Force Research Laboratory (AFRL), and awarded to
Raytheon BBN Technologies under contract number FA865018-C-7885.
Any opinions, findings, conclusions, or recommendations
expressed here are those of the authors and do not
necessarily reflect the view of the sponsors.

\putbib[bib/journal-full,bib/ref]
\end{bibunit}

\clearpage
\appendix

\begin{bibunit}[acl_natbib_2019]

\appendix

\setcounter{table}{0}
\renewcommand{\thetable}{\Alph{section}\arabic{table}}

\section{Model Implementation Details}
\label{sec:app-impl}
We use and compare two different feature extractors in our experiments.
For bi-LSTM encoders, we concatenate $100$-dimensional word embeddings initialized from pre-trained \glove{} embeddings \citep{pennington+14},
$8$-dimensional part-of-speech and $8$-dimensional named-entity embeddings as input to the LSTM encoders.
Tokens that appear less than five times
are replaced with a special ``\abr{unk}'' token.
For the BERT setting, we fine-tune a \abr{BERT}\textsubscript{base} model\footnote{\url{https://github.com/huggingface/transformers}}
and use the $768$-dimensional final-layer representations.
For the decoder,
we embed previously decoded tokens, such as keywords,
into $256$-dimensional vectors and feed them as next-timestep input to the decoder LSTM.
Both the encoder and decoder LSTMs
have $128$ hidden units and $2$ layers.
If the decoder predicts question words as literal strings in the output \abr{sql} queries,
we replace them with the most similar table cell values using
fuzzy match.\footnote{\url{https://github.com/seatgeek/fuzzywuzzy}}
We set both $\lambda^\text{att}$ and $\lambda^\text{CP}$ to be $0.2$.
During training, we use a batch size of $8$ and we set the dropout rate to be $0.3$ in all MLPs and LSTMs. %
We use the Adam optimizer \cite{kingma2014adam} with default learning rate $0.001$
and we clip gradients to $5.0$. %
We train our models for up to $50$ epochs and conduct early stopping based on per-epoch dev-set evaluation.
On a single GTX 1080 Ti GPU, a training mini-batch takes $0.7$ second on average
and the training process finishes within $10$ hours.
We do not tune hyper-parameters.

\section{Comparison of Our Baseline Model with a State-of-the-Art Text-to-\abr{sql} Parser}
\label{sec:app-sota}
To evaluate the strength of our baseline model,
we compare it with \posscite{alane2020explor} state-of-the-art model previously tested on the Spider dataset \citep{yu2018spider}.
Our task formulation is unlike the Spider dataset in that
1) the official Spider evaluation does not require predictions of literal values
and 2) on our dataset, the model needs to predict data types for each column (e.g., \sql{\_number} in \reffig{fig:ex}).
\posscite{alane2020explor} model has already addressed the first difference by including literal string prediction modules,
and we loosen our evaluation criteria for the sake of this comparison.
We train \posscite{alane2020explor} model on \wonderdata with their reported hyperparameters
and evaluate with a variant of logical form accuracies (\acclfminus)
that accepts column type disparities between the prediction and the gold standard;
\reftab{tb:stoa-compare} shows the evaluation results.
Our baseline \baseline{} model has competitive \acclfminus with \posscite{alane2020explor} state-of-the-art text-to-\abr{sql} parser.

\begin{table}[t]
    \centering
    \footnotesize
    \begin{tabular}{lc}

      \toprule
       Model & \acclfminus
       \\
       \midrule
       \baselinebert{}   & $50.8$  \\
       \citet{alane2020explor} w/ \abr{bert}   & $51.7$  \\
      \bottomrule
    \end{tabular}
      \caption{
      Dev logical form accuracy excluding column type (\acclfminus) of our \baselinebert{}
      is comparable to that of
      a state-of-the-art model on Spider.%
      }
    \label{tb:stoa-compare}
  \end{table}

\section{Annotation Guidelines}
\label{sec:app-anno}

In our pilot study, we instruct two expert \abr{sql} annotators to write down \abr{sql} equivalents of the English questions
and to pick out the lexical mappings between the question and \abr{sql} tokens
that correspond to each other semantically and are atomic, i.e., they cannot further decompose into smaller meaningful mappings.
These underspecified instructions lead to $70.4\%$ agreement on \abr{sql} annotation
and $75.1\%$ agreement on alignment annotation.
The annotators have similar but not identical intuitions about, for example, what constitutes an atomic unit,
especially when there are equally plausible alternative options.
Following discussions,
we refine our annotation guidelines for frequently occurring patterns
to ensure consistent annotations, as follows:

\subsection*{\centering General Rules}
\begin{enumerate*}
    \item \abr{sql} queries should reflect the semantic intent of the English questions,
    even if shorter \abr{sql} queries return the same execution results.
    The only exception is when \abr{sql} offers no straightforward implementation of the implicit semantic constraints.
    In that case,
    answer the first appearing subquestion, i.e., assume that the implicit semantic constraints are always met.
    For example, it is implicitly assumed in the question ``which city are A and B located in?''
    that A and B are located in the same city;
    write down the \abr{sql} equivalent for ``which city is A located in?''.
    \item When there are competing choices of annotation, select the simplest version.
    Among alternative \abr{sql} queries, select the one with fewer nestings and fewer \abr{sql} tokens:
    \sql{SELECT MAX(col) FROM w} is prioritized over
    \sql{SELECT col FROM w ORDER BY col DESC LIMIT 1}.
    Following this rule, default values are always omitted since the queries are shorter without them.
    These include, for example,
    the keyword \sql{ASC} in an \sql{ORDER BY} clause.
    \item Lexical alignments should cover as many semantically-meaningful tokens as possible,
    even if there is no word overlap.
    For example, for the question ``who performed better, toshida or young-sun?'',
    align the word ``performed'' to its corresponding column (``result'' or ``rank'').
    For \emph{wh}-tokens, align ``when'', ``who'' and ``where'' if appropriate,
    but omit alignments of ``what'' and ``which''
    when they do not contribute to concrete meanings.
    \item Prioritize alignments with exact lexical matches.
    This means that for many noun phrases,
    align bare nouns excluding the determiners instead of maximal noun phrases
    (e.g., ``movie'' rather than ``the movie'' should be aligned to the ``movie'' column token in the \abr{sql} query).
    In contrast, include ``the'' in the alignment of superlatives (e.g., ``the least''),
    since superlatives usually do not lexically overlap with the column tokens.
    \item In general, the annotation should not depend on the table contents and sorting assumptions.
    In other words, use direct references to the presented row order \sql{id} as little as possible.
    However, use \sql{id} if the question explicitly asks about the presentation order,
    e.g., ``the first on the list'' or ``the first listed''.
\end{enumerate*}

\subsection*{\centering Some Frequent Specific Cases}

\begin{enumerate*}
    \item Align ``how many'' to the aggregation operation when appropriate,
    but do not align ``how many''
    when the \abr{sql} query directly selects a column without aggregation,
    e.g., the question is ``how many total medals has Spain won?'' and the table contains a column ``total''.
    \item Only add the keyword \sql{DISTINCT} if there are clear linguistic cues
    (``how many \emph{different} countries on the table?''),
    otherwise do not use \sql{DISTINCT}.
    \item Use \sql{COUNT(col)} if possible and use \sql{COUNT(*)}
    only if there is no good match from the question to any column.
    \item When the question asks about the row with the max/min value in a column,
    generally use \sql{SELECT col FROM w ORDER BY col [DESC] LIMIT 1}.
    If there are ties in the max/min values, use \sql{SELECT col FROM w WHERE col = (SELECT MAX(col) FROM w)}.
    \item Align question word ``game'' to ``date'' column if necessary
    but use \sql{COUNT(*)} for counting the game numbers
    when there are no better alignment alternatives.
    \item Align words referring to performance, such as ``fast'', to the corresponding ``result''/``time'' columns;
    if not available, align them to ``rank'' columns that indirectly refer to performance;
    if still not available, align them to \sql{id}, which explicitly relies on the table being presorted by the performance.
\end{enumerate*}

\section{Database Construction}
\label{sec:app-database}

We assume $9$ basic data types for \wtq tables:
numbers (e.g., ``$5$''), numbers with units (e.g., ``$5$ kg'') , date and time (e.g., ``May $29$, $1968$'', ``$3$:$56$''), (sports) scores (e.g., ``w $5$:$3$''),
number spans (e.g., ``$12$--$89$''), time spans (e.g., ``May $2011$--June $2012$''), fractions (e.g., ``$3$/$5$'') , street addresses (e.g., ``$2020$ Westchester Street''), and raw texts (e.g., ``John Shermer'').
Additionally, we consider two composite types: binary tuples (e.g., ``KO (head kick)'') and lists (e.g., ``Wojtek Fibak,
Joakim Nyström'').
Binary tuples are split into two sub-columns in the generated databases,
and lists are automatically transformed to a separate table joined with the original table
through primary-foreign key relations.
Data types for each column are first identified with regular expressions
and manually verified by
annotators.
Any column that contains a type outside of these $9$ types
is interpreted as
raw text.
We also filter out aggregation rows from the tables
so that the \abr{sql} aggregation functions over the table can skip those pre-computed aggregates.

\section{Additional Alignment Data Statistics}
\label{sec:app-pos-tag}

\begin{table*}[tbp]
  \centering
    \small
    \begin{tabular}{rr|rr}
    \toprule
    POS & (\%)$\downarrow$ & POS & (\%)$\uparrow$ \\
    \midrule
  \begin{tabular}{@{\hspace{0pt}}r@{\hspace{0pt}}}RBS (Adverb, superlative)\end{tabular} & $99.02$ & \begin{tabular}{@{\hspace{0pt}}r@{\hspace{0pt}}}. (Punctuation)\end{tabular}       &  $0.15$ \\
  \begin{tabular}{@{\hspace{0pt}}r@{\hspace{0pt}}}JJR (Adjective, comparative)\end{tabular} & $96.24$ &\begin{tabular}{@{\hspace{0pt}}r@{\hspace{0pt}}}WDT (\emph{wh}-determiner)\end{tabular}  &  $1.20$ \\
  \begin{tabular}{@{\hspace{0pt}}r@{\hspace{0pt}}}JJS (Adjective, superlative)\end{tabular} & $94.66$ & \begin{tabular}{@{\hspace{0pt}}r@{\hspace{0pt}}}VBD-AUX (Auxiliary verb)\end{tabular}  &  $2.26$ \\
  \begin{tabular}{@{\hspace{0pt}}r@{\hspace{0pt}}}RBR (Adverb, comparative)\end{tabular}  & $93.89$ & \begin{tabular}{@{\hspace{0pt}}r@{\hspace{0pt}}}EX (Existential \emph{there})\end{tabular}       &  $3.56$ \\
  \begin{tabular}{@{\hspace{0pt}}r@{\hspace{0pt}}}WRB (\emph{wh}-adverb)\end{tabular}  & $88.25$ & \begin{tabular}{@{\hspace{0pt}}r@{\hspace{0pt}}}PRP\$ (Possessive pronoun)\end{tabular}    &  $9.38$ \\
  \begin{tabular}{@{\hspace{0pt}}r@{\hspace{0pt}}}JJ (Adjective)\end{tabular}   & $82.07$ & \begin{tabular}{@{\hspace{0pt}}r@{\hspace{0pt}}}POS (Possessive ending)\end{tabular}      & $13.42$ \\
  \begin{tabular}{@{\hspace{0pt}}r@{\hspace{0pt}}}CD (Cardinal number)\end{tabular}   & $79.48$ & \begin{tabular}{@{\hspace{0pt}}r@{\hspace{0pt}}}PRP (Personal pronoun)\end{tabular}      & $13.95$ \\
  \begin{tabular}{@{\hspace{0pt}}r@{\hspace{0pt}}}NNP (Proper noun, singular)\end{tabular}  & $75.70$ & \begin{tabular}{@{\hspace{0pt}}r@{\hspace{0pt}}}WP (\emph{wh}-pronoun)\end{tabular}       & $20.58$ \\
    \bottomrule
    \end{tabular}%
    \caption{
        The POS tags with the highest and lowest alignment ratios (\%) to \abr{sql} queries
        (with more than $100$ occurrences).
    Comparative/superlative
   adjectives (JJR, JJS) and adverbs (RBS, RBR) are most aligned,
    corresponding to \abr{sql} operations like \sql{MAX}.
    Punctuations (.),
    \emph{wh}-determiners (WDT),
    helper-verbs (VBD-AUX),
    existential \emph{there}'s (EX),
    and pronouns (PRP, PRP\$) are least
   aligned.
    }

  \label{tab:pos-stats}%

\end{table*}%

\reftab{tab:pos-stats} shows the part-of-speech tags that are most- and least-aligned.\footnote{
These POS tags are automatically derived from Stanford CoreNLP toolkit
and are provided in the \wtq dataset.
}
Comparative and superlative adjectives and adverbs are among the most frequently aligned tokens,
while pronouns and function words are
infrequently aligned.

\section{Different Loss Functions for Supervised Attention}
\label{sec:app-abl-loss}

Following \citet{liu+16}, we experiment with three different attention loss definitions:
\begin{align}
    L^{\text{att}}&=
    \frac{1}{2}\lVert\mathbf{a} - \mathbf{a}^\star\rVert^2\tag{Mean Squared Error}\\
    L^{\text{att}}&=
    -\log\left(\mathbf{a} \cdot \mathbf{a}^\star\right)\tag{Multiplication}\\
    L^{\text{att}}&=
     -\mathbf{a}^\star \cdot \log\left(\mathbf{a}\right),\tag{Cross Entropy}
\end{align}
where $\mathbf{a}_i$ and $\mathbf{a}_i^\star$ denote the learned attention weights
and annotated gold-standard alignments.
A smaller distance between
$\mathbf{a}_i$ and $\mathbf{a}_i^\star$
indicates a model better at reproducing our alignment annotation.
While both mean squared error and multiplication are symmetric in $\mathbf{a}_i$ and $\mathbf{a}_i^*$,
cross entropy is asymmetric and has been previously shown to be the most effective measure in the task of machine translation \citep{liu+16}.
Table~\ref{tb:attn-loss}
shows dev-set results with different supervised attention loss choices in \alignmodel{}'s encoder.
The mean square error loss is the strongest, with
$1.5\%$ higher execution accuracy than multiplication
loss and $0.6\%$ higher than
cross-entropy loss.

\begin{table}[t]
  \centering
  \footnotesize
  \begin{tabular}{l|cc}

    \toprule
     Attention Loss & \acclf & \accexe \\
     \midrule
     Mean squared error (\alignmodel{})    & $41.8\pm1.6$&$60.9\pm0.8$  \\
     Multiplication   & $40.3\pm1.5$ & $59.4\pm1.0$ \\
     Cross entropy     & $41.6\pm1.2$ & $60.3\pm1.0$  \\
    \bottomrule
  \end{tabular}
    \caption{
    Dev logical form (\acclf) and execution (\accexe) accuracies with different attention loss functions.
    Our final model \alignmodel{} uses mean squared error,
    the most accurate variant of the three loss functions.
    }
  \label{tb:attn-loss}
\end{table}

\section{\alignmodel Trained with Heuristically-Generated Alignments}

\begin{table}[t]
  \centering
  \footnotesize
  \begin{tabular}{lcc}
    \toprule
    \multirow{2}{*}{Model} & \multicolumn{2}{c}{Dev}\\
    & \acclf & \accexe \\
    \midrule
    \baseline  & $37.8\pm0.6$  & $56.9\pm0.7$ \\
    \alignmodel (Heuristics)    & $40.3\pm1.8$  & $59.6\pm1.4$ \\
    \alignmodel (Manual)  & $42.2\pm1.5$& $61.3\pm0.8$\\
      \midrule
  \end{tabular}
    \caption{
    Dev logical form (\acclf) and execution (\accexe) accuracies
    comparing \alignmodel trained with automatic and manual alignments.
    Training with automatic alignments leads to higher accuracies
    than \baseline and manual annotations give an additional accuracy improvement.
    }
  \label{tb:addt-ablation}
\end{table}

We experiment with question-column alignments
derived from textual fuzzy matching between column names and question $5$-grams.
\reftab{tb:addt-ablation} shows dev-set results.
Training with automatic alignments improves over the \baseline model,
but manual annotations provide an additional $+1.7\%$ \accexe.
The manual annotations are cleaner and more informative
since there are many column mentions without any lexical overlap with the column headers
(e.g., ``who'' $\leftrightarrow$ column ``athlete'').

\section{Template-based Evaluation}
\label{sec:app-temp}

\begin{table*}[h]
  \centering
  \small
  \begin{tabular}{l@{\hspace{3pt}}r@{\hspace{6pt}}c@{\hspace{3pt}}c@{\hspace{3pt}}c@{\hspace{3pt}}c@{\hspace{3pt}}c@{\hspace{3pt}}c}
    \toprule
    \multirow{2}{*}{Template} & \multirow{2}{*}{Count} & \multicolumn{2}{c}{\acclf}& \multicolumn{2}{c}{\accskt}& \multicolumn{2}{c}{\acccol}\\
     & &\baseline{}&\alignmodel{}&\baseline{}&\alignmodel{}&\baseline{}&\alignmodel{}\\
    \midrule
    \sql{SELECT col FROM w ORDER BY} & \multirow{2}{*}{$1{,}490$} & \multirow{2}{*}{$48.1$}&\multirow{2}{*}{$\mathbf{50.6}$}&\multirow{2}{*}{$86.9$}&\multirow{2}{*}{$\mathbf{87.6}$}&\multirow{2}{*}{$56.3$}&\multirow{2}{*}{$\mathbf{60.2}$}\\
    \quad\sql{col [DESC] LIMIT 1} &\\[5pt]
    \sql{SELECT col FROM w WHERE col = STR} & \multirow{1}{*}{$1{,}149$} & \multirow{1}{*}{$39.5$}&\multirow{1}{*}{$\mathbf{42.6}$}&\multirow{1}{*}{$73.6$}&\multirow{1}{*}{$\mathbf{75.0}$}&\multirow{1}{*}{$40.1$}&\multirow{1}{*}{$\mathbf{44.0}$}\\[5pt]
    \sql{SELECT COUNT(col) FROM w WHERE col = STR} & \multirow{1}{*}{$1{,}127$} & \multirow{1}{*}{$55.0$}&\multirow{1}{*}{$\mathbf{59.8}$}&\multirow{1}{*}{$85.2$}&\multirow{1}{*}{$\mathbf{86.1}$}&\multirow{1}{*}{$55.9$}&\multirow{1}{*}{$\mathbf{60.3}$}\\[5pt]
    \sql{SELECT COUNT(col) FROM w WHERE col COMP NUM} & \multirow{1}{*}{$635$} & \multirow{1}{*}{$50.1$}&\multirow{1}{*}{$\mathbf{57.6}$}&\multirow{1}{*}{$89.0$}&\multirow{1}{*}{$\mathbf{91.1}$}&\multirow{1}{*}{$57.8$}&\multirow{1}{*}{$\mathbf{66.0}$}\\[5pt]
    \sql{SELECT col FROM w WHERE col = NUM} & \multirow{1}{*}{$607$}  & \multirow{1}{*}{$49.4$}&\multirow{1}{*}{$\mathbf{54.7}$}&\multirow{1}{*}{$72.9$}&\multirow{1}{*}{$\mathbf{75.3}$}&\multirow{1}{*}{$49.7$}&\multirow{1}{*}{$\mathbf{55.0}$}\\[5pt]
    \sql{SELECT COUNT(col) FROM w} & \multirow{1}{*}{$507$} & \multirow{1}{*}{$43.2$}&\multirow{1}{*}{$\mathbf{51.3}$}&\multirow{1}{*}{$\mathbf{78.1}$}&\multirow{1}{*}{$77.7$}&\multirow{1}{*}{$48.9$}&\multirow{1}{*}{$\mathbf{59.4}$}\\[5pt]
    \sql{SELECT col FROM w GROUP BY col ORDER BY} & \multirow{2}{*}{$315$} & \multirow{2}{*}{$34.6$}&\multirow{2}{*}{$\mathbf{47.3}$}&\multirow{2}{*}{$80.0$}&\multirow{2}{*}{$\mathbf{85.4}$}&\multirow{2}{*}{$36.2$}&\multirow{2}{*}{$\mathbf{49.5}$}\\
    \quad\sql{COUNT(col) [DESC] LIMIT 1}  &\\[5pt]
    \sql{SELECT COUNT(col) FROM w WHERE col = NUM} & \multirow{1}{*}{$308$} & \multirow{1}{*}{$51.0$}&\multirow{1}{*}{$\mathbf{59.8}$}&\multirow{1}{*}{$85.1$}&\multirow{1}{*}{$\mathbf{87.3}$}&\multirow{1}{*}{$51.9$}&\multirow{1}{*}{$\mathbf{59.7}$}\\[5pt]
    \sql{SELECT col FROM w WHERE col = (SELECT} & \multirow{2}{*}{$284$} & \multirow{2}{*}{$61.2$}&\multirow{2}{*}{$\mathbf{61.6}$}&\multirow{2}{*}{$\mathbf{76.1}$}&\multirow{2}{*}{$75.7$}&\multirow{2}{*}{$61.6$}&\multirow{2}{*}{$\mathbf{62.0}$}\\
    \quad\sql{col FROM w WHERE col = STR) + 1} &\\[5pt]
    \sql{SELECT col FROM w WHERE col IN (STR, STR)} & \multirow{2}{*}{$282$} & \multirow{2}{*}{$39.0$}&\multirow{2}{*}{$\mathbf{46.8}$}&\multirow{2}{*}{$85.5$}&\multirow{2}{*}{$\mathbf{85.8}$}&\multirow{2}{*}{$49.3$}&\multirow{2}{*}{$\mathbf{56.0}$}\\
    \quad\sql{ORDER BY col [DESC] LIMIT 1} &\\
    \midrule
    \emph{Entire Corpus} & $11{,}276$ & $37.8$ & $\mathbf{42.2}$ & $64.7$ & $\mathbf{66.7}$ & $39.6$ & $\mathbf{44.5}$ \\
    \bottomrule
\end{tabular}
    \caption{
      Dev logical form (\acclf), template (\accskt) and column (\acccol) accuracies
    on the $10$ most frequent templates.
     We combine model predictions from five data splits for this analysis.
     \sql{[DESC]} denotes the keyword \sql{DESC} is optional,
     and \sql{COMP} includes comparison operators ($>$, $<$, $>=$, $<=$ and $\neq$).
     \alignmodel yields higher \acccol gains on complex templates,
     compared with simple and common templates.
    }
  \label{tb:temp-result-full}
\end{table*}

\addtocounter{section}{1}
\begin{table*}[!h]
    \centering
    \small
    \begin{tabular}{lrcccc}
      \toprule
      \multirow{2}{*}{Unseen Template} & \multirow{2}{*}{Count} & \multicolumn{2}{c}{\acclf}& \multicolumn{2}{c}{\accexe}\\
      & &\baseline{}&\alignmodel{}&\baseline{}&\alignmodel{}\\
      \midrule
      \sql{SELECT col FROM w ORDER BY}   & \multirow{2}{*}{$1{,}490$} & \multirow{2}{*}{~~$9.0$}&\multirow{2}{*}{$\mathbf{23.1}$}&\multirow{2}{*}{$38.9$}&\multirow{2}{*}{$\mathbf{48.2}$}\\
      \quad\sql{col [DESC] LIMIT 1} &\\[5pt]
      \sql{SELECT col FROM w WHERE col = STR} & $1{,}149$  & \multirow{1}{*}{$\mathbf{12.8}$}&\multirow{1}{*}{$11.3$}&\multirow{1}{*}{$48.8$}&\multirow{1}{*}{$\mathbf{53.7}$}\\[5pt]
      \sql{SELECT COUNT(col) FROM w WHERE col = STR} & $1{,}127$ & \multirow{1}{*}{~~$9.0$}&\multirow{1}{*}{$\mathbf{34.0}$}&\multirow{1}{*}{$32.0$}&\multirow{1}{*}{$\mathbf{57.0}$}\\[5pt]
      \sql{SELECT COUNT(col) FROM w WHERE col COMP NUM} & $635$ & \multirow{1}{*}{$22.6$}&\multirow{1}{*}{$\mathbf{45.2}$}&\multirow{1}{*}{$51.6$}&\multirow{1}{*}{$\mathbf{58.9}$}\\[5pt]
      \sql{SELECT col FROM w WHERE col = NUM} & $607$  & \multirow{1}{*}{$15.4$}&\multirow{1}{*}{$\mathbf{19.5}$}&\multirow{1}{*}{$58.5$}&\multirow{1}{*}{$\mathbf{68.3}$}\\[5pt]
      \sql{SELECT COUNT(col) FROM w} & $507$  & \multirow{1}{*}{~~$0.0$}&\multirow{1}{*}{~~$\mathbf{1.0}$}&\multirow{1}{*}{$19.0$}&\multirow{1}{*}{$\mathbf{23.0}$}\\[5pt]
      \sql{SELECT col FROM w GROUP BY col ORDER BY} & \multirow{2}{*}{$315$} & \multirow{2}{*}{~~$3.3$}&\multirow{2}{*}{$\mathbf{50.8}$}&\multirow{2}{*}{$24.6$}&\multirow{2}{*}{$\mathbf{73.8}$}\\
      \quad\sql{COUNT(col) [DESC] LIMIT 1} &\\[5pt]
      \sql{SELECT COUNT(col) FROM w WHERE col = NUM} & $308$ & \multirow{1}{*}{$\mathbf{34.0}$}&\multirow{1}{*}{$30.0$}&\multirow{1}{*}{$59.0$}&\multirow{1}{*}{$\mathbf{66.0}$}\\[5pt]
      \sql{SELECT col FROM w WHERE col = (SELECT} & \multirow{2}{*}{$284$} & \multirow{2}{*}{$\mathbf{30.8}$}&\multirow{2}{*}{$15.4$}&\multirow{2}{*}{$\mathbf{61.5}$}&\multirow{2}{*}{$57.7$}\\
      \quad\sql{col FROM w WHERE col = STR) + 1} &\\[5pt]
      \sql{SELECT col FROM w WHERE col IN (STR, STR)} & \multirow{2}{*}{$282$} & \multirow{2}{*}{$17.9$}&\multirow{2}{*}{$\mathbf{30.4}$}&\multirow{2}{*}{$53.6$}&\multirow{2}{*}{$\mathbf{66.4}$}\\
      \quad\sql{ORDER BY col [DESC] LIMIT 1} &\\
      \midrule
      \emph{Macro-average over the above templates} & --- & $15.5$ & $\mathbf{26.1}$ & $44.8$ & $\mathbf{57.3}$ \\
      \bottomrule
  \end{tabular}
      \caption{
        Dev logical form (\acclf) and execution (\accexe) accuracies in a generalization evaluation setting
        following \citet{finegan+2018},
        where instances of a given template are ablated from training,
        and we evaluate model accuracies on that unseen template.
      \alignmodel outperforms \baseline in \accexe on $9$ out of the $10$ most frequent templates.
      }
    \label{tb:temp-gene-result-full}

  \end{table*}
\addtocounter{section}{-1}

\reftab{tb:temp-result-full} shows dev-set results of the top $10$ most frequent templates.
We report logical form (\acclf), template (\accskt) and column (\acccol) accuracies.
\acccol is calculated on the subset where template predictions are accurate.\footnote{
We do not include literal string and number accuracies:
both \baseline and \alignmodel get nearly perfect scores ($>98\%$).
}
The improvement of \alignmodel over \baseline is more significant on \acccol than \accskt.
Additionally, \alignmodel tends to yield higher \acccol gains
on complex templates,
compared with simple and common templates.
\section{Evaluation Results on Unseen \abr{sql} Templates}
\label{sec:app-generalization}

\reftab{tb:temp-gene-result-full} considers an evaluation setting of \citet{finegan+2018}
to test the model accuracies on unseen \abr{sql} templates.
We exclude all instances of a given template from the training set,
and then evaluate only on that template.
\alignmodel{} outperforms \baseline{} in  \accexe on $9$ out of the $10$ most frequent templates.
Notably, on a template that contains both \sql{GROUP BY} and \sql{ORDER BY} clauses,
the \accexe improvement of \alignmodel is as large as $+49.2\%$.

\putbib[bib/journal-full,bib/ref]
\end{bibunit}

\end{document}